\documentclass{article}
\usepackage{spconf,amsmath,graphicx, multirow}
\usepackage{tikz}
\usepackage[boldmath]{numprint}
\usepackage{siunitx}
\usepackage{tablefootnote}
\usepackage[]{algorithm2e}
\usepackage[backend=biber, style=ieee, maxnames=4, doi=false,isbn=false,url=false,citestyle=numeric-comp]{biblatex}
\DeclareSIUnit{\microsecond}{\SIUnitSymbolMicro s}

\usepackage{footnote}
\makesavenoteenv{tabular}

\usepackage{pgfplots}
\pgfplotsset{compat=1.14}

\def\x{{\mathbf x}}
\def\L{{\cal L}}

\title{GPU-Accelerated Viterbi Exact Lattice Decoder for\\Batched Online and Offline Speech Recognition}
\name{Hugo Braun$^{\dagger}$, Justin Luitjens$^{\dagger}$, Ryan Leary$^{\dagger}$, Tim Kaldewey$^{\dagger}$, Daniel Povey}

\address{
 $^{\dagger}$NVIDIA, Santa Clara, USA
}

\begin{document}
\maketitle{}
\begin{abstract}
We present an optimized weighted finite-state transducer (WFST) decoder capable of online streaming and offline batch processing of audio using Graphics Processing Units (GPUs).
The decoder is efficient in memory utilization, input/output (I/O) bandwidth, and uses a novel Viterbi implementation designed to maximize parallelism.
The reduced memory footprint allows the decoder to process significantly larger graphs than previously possible, while optimizing I/O increases the number of simultaneous streams supported. 
GPU preprocessing of lattice segments enables intermediate lattice results to be returned to the requestor during streaming inference. 
Collectively, the proposed algorithm yields up to a 240x speedup over single core CPU decoding, and up to 40x faster decoding than the current state-of-the-art GPU decoder, while returning equivalent results.
This decoder design enables deployment of production-grade ASR models on a large spectrum of systems, ranging from large data center servers to low-power edge devices.
\end{abstract}
\section{Introduction}
\label{sec:intro}
Recent advancements in automatic speech recognition (ASR), fueled by deep learning research in the field \cite{hinton}, have led to significant quality improvements, making the technology practical for a slew of human-computer interaction use cases and driving demand for streaming ASR as a service.
Streaming ASR as a service typically requires large numbers of commodity servers in a datacenter. 
Tight latency requirements guided work to improve inference performance of models deployed in datacenters and encouraged research on supporting inference at the edge, including low-power devices \cite{he, mcgraw}.

Typical ASR systems comprise three primary components: feature extraction, acoustic modeling, and language model decoding. 
Historically, the computational complexity of the acoustic model has dominated the inference execution time, and has been the focus of a variety of optimizations, including unusual network architectures, striding, and quantization techniques \cite{pundak2016, xiang2017, peddinti2018}.

Principal among these optimizations is offloading acoustic model inference to dedicated acceleration hardware, most commonly GPUs \cite{dixon2009}.
In many cases, feature extraction and neural acoustic models are efficient enough such that further optimization is limited by Amdahl's law \cite{rodgers1985}: marginal latency improvements in previously optimized components yield negligible improvements in system latency. 
To begin our investigation into accelerating speech recognition inference, we profiled a typical lattice decode using the Kaldi speech recognition framework \cite{Povey_ASRU2011} with a pretrained model (see experiments in Section \ref{sec:experiments}), and found 94\% of the wallclock time was spent in the language model decoder when using a GPU for acoustic model inference.

In this work, we propose a novel implementation of weighted finite-state transducer (WFST) decoding for the speech recognition task using GPUs and NVIDIA's CUDA \cite{cuda} programming language. 
The decoder is designed as a drop-in replacement for existing decoders, requiring no language or acoustic model modifications.
It is designed to be maximally flexible, supporting online recognition of multiple simultaneous audio streams and lattice generation.
Carefully bounded memory utilization ensures adequate space on GPU memory for large language models and coresident acoustic models.
Finally, the algorithm can scale from small GPUs running on low-power embedded GPUs to multiple datacenter-class GPUs running in a single server.
Prior to publication, the work has been open-sourced and is now included with Kaldi\footnote{https://github.com/kaldi-asr/kaldi/tree/master/src/cudadecoder}.

\section{Related Work}
\label{sec:related}
Originally proposed by Mohri \cite{mohri2002}, WFSTs for ASR decoding have become the de facto standard when using n-gram language models. 
The decode process returns the single-best path, or alternatively an exact lattice \cite{povey2012} representing multiple possible hypotheses for the decoded utterance.
Efforts to increase the speed of the decode and lattice generation process have included parallel, multi-threaded CPU implementations \cite{mendis2016} as well as hybrid on-the-fly rescoring \cite{hori2004}.

Despite promising efforts in \cite{mendis2016}, attempts to extend previous accelerated speech decoding onto parallel processors are relatively nascent.
Initial efforts targeted hybrid rescoring methods \cite{sak2010, kim2012} using constrained vocabularies or language models on GPU, while offloading rescoring to CPU.
General-purpose WFST decoding on GPUs has been proposed in \cite{zhang2009, hanif2017, argueta2017}, but these works do not support conditioning on acoustic model (AM) posteriors.

The proposed work is most closely related to and improves upon the first fully GPU-accelerated lattice decoder \cite{Chen2018}, which maps token passing constructs \cite{mendis2016} to GPU.
Starting from the single-threaded CPU decoder, we tailored the algorithm to the strengths of the hardware, including avoiding unnecessary synchronization and atomics, and using flat, compact memory structures. Efficiencies realized in this implementation enabled the addition of support for online decoding while achieving up to 40x speedups over previous accelerated implementations.

\section{Parallel Viterbi Decoding}
\label{sec:viterbi}
The parallel WFST decoder generally follows the typical order of operations in a serial decoder: for each frame of AM posteriors, the decoder processes emitting arcs (those arcs with non-null labels) conditioned on frame values, processes any chains of non-emitting arcs, and finally performs pruning. 
The proposed algorithm utilizes two disparate asynchronous CUDA streams: one responsible for executing compute kernels, and the other responsible for performing non-blocking device to host (D2H) memory copies of lattice tokens.
Using a second stream for D2H copies makes it possible to return intermediate results during online coding without stalling the compute pipeline.

We eliminate many common CPU-oriented optimizations and constraints, which are sometimes detrimental to parallel performance. Specifically, when expanding tokens, we do not test that new tokens are unique. It is sufficient for correctness to allow duplicate tokens to persist and be cleaned later: trading marginal extra work for reduced dependence on synchronization and atomic operations.
Despite further micro-optimizations in the code, we focus this section on the unique architectural decisions of the decoder for brevity.

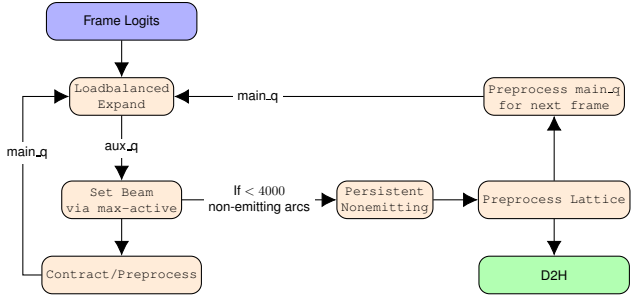
\begin{figure}
\usetikzlibrary{arrows, arrows.meta}
\tikzset{%
  >={Latex[width=2mm,length=2mm]},
            base/.style = {rectangle, rounded corners, draw=black,
                           minimum width=4cm, minimum height=1cm,
                           text centered, font=\sffamily, scale=0.5},
  activityStarts/.style = {base, fill=blue!30},
       startstop/.style = {base, fill=red!30},
    d2h/.style = {base, fill=green!30},
         kernel/.style = {base, minimum width=2.5cm, fill=orange!15,
                           font=\ttfamily},
         label/.style = {scale=0.5}
}

\begin{tikzpicture}[node distance=1.5cm,
    every node/.style={fill=white, font=\sffamily}, align=center, scale=0.5]
  \node (start)             [activityStarts]{Frame Logits};
  \node (lbExpand)     [kernel, below of=start,yshift=-0.5cm]{Loadbalanced\\Expand};
  \node (setBeam)      [kernel, below of=lbExpand, yshift=-1.25cm]{Set Beam\\via max-active};
  \node (contractPreproc)     [kernel, below of=setBeam, yshift=-0.5cm]{Contract/Preprocess};
  \node (persistNE)    [kernel, right of=setBeam, xshift=5.5cm]{Persistent\\Nonemitting};
  \node (genLattice)      [kernel, right of=persistNE, xshift=3cm]{Preprocess Lattice};
  \node (copyD2H)      [d2h, below of=genLattice, yshift=-0.5cm]{D2H};
  \node (preprocNF)      [kernel, above of=genLattice, yshift=1.25cm]{Preprocess main\_q\\for next frame};
  \draw[->]             (start) -- (lbExpand);
  \draw[->]     (lbExpand) -- node[label]{aux\_q}(setBeam);
  \draw[->]      (setBeam) -- (contractPreproc);
  \draw[->]    (setBeam) -- node[label]{If $< 4000$\\non-emitting arcs}(persistNE);
  \draw[->]    (persistNE) -- (genLattice);
  \draw[->]    (genLattice) -- (copyD2H);
  \draw[->]    (genLattice) -- (preprocNF);
  \draw[->]    (contractPreproc.west) -- ++(-0.5,0) -- ++(0,4.75) -- node[label,yshift=-1.5cm, xshift=-0.5cm]{main\_q}(lbExpand.west);
  \draw[->]    (preprocNF.west) |- node[label,xshift=-6cm]{main\_q}(lbExpand);
\end{tikzpicture}
\caption{Block diagram of kernels involved in advancing decoding.}
\label{fig:adv_dec}
\end{figure}

\subsection{Batching \& Context Switching}
As decoding is necessarily serial in nature (i.e. prediction at time $t$ depends on the state at $t-1$), and individual steps represent relatively small units of work, decoding functions (kernels) executing on the GPU complete quickly, and performance becomes constrained by kernel launch latency. By structuring the decoder such that multiple audio streams are processed in parallel, launch latency is hidden by longer-running kernels (due to their increased workload).

To support efficient decoding for online recognition, we introduce two separate mechanisms for handling simultaneous audio streams: channels and lanes. Lanes are roughly equivalent to batch size in neural networks, and represent the set of utterances or streams being actively decoded. Channels maintain state for utterances which are not ready to continue processing due to lack of audio or computed posteriors. The threaded decoder that readies work for the GPU is responsible for multiplexing channels (as they become ready) onto lanes (as they become available). This scheme allows for easy tuning to match the GPU with the model and representative data: increase the number of lanes until diminishing returns are reached, and set the number of channels to match the measured throughput/xRTF.

Critical to this strategy is the ability to efficiently swap channels with lanes, which requires minimizing memory usage required for state tracking and optimizing layout. In practice, context switching calls complete in about 5\si{\microsecond} per batch. Details of the memory structure used is described in the following section.

\subsection{Memory Layout}
\label{ssec:mem}
Maximum efficiency depends on minimizing memory usage for state. Equally important is the layout of memory. Careful consideration is taken here to ensure that data is structured such that kernels may use coalesced accesses wherever possible. 

\subsubsection{Footprint}
We represent the decoding FST in-memory as a set of compressed sparse rows (CSRs) and additional metadata, which we are able to efficiently traverse with direct indexing.

Given the decode WFST $T = (\Sigma, \Omega, Q, E, ...)$, with input and output labels $\Sigma$ and $\Omega$, respectively, a finite set of states $Q$, a finite set of transitions $E$ ($E_E$ are emitting transitions), we calculate its expected memory utilization, $M_{fst}$ as
\begin{equation}
    M_{fst} = 12 |Q| + 8 |E| +  4 |E_E|
\end{equation}
In practice, this typically equates to GPU memory used for the FST about $\frac{1}{3}$ of the size of the FST on disk.

GPU memory utilization of the decoder is bounded and can be calculated with a closed-form equation based on configured hyperparameters.
The memory footprint, in bytes, of the full state of the decoder, including utterances being actively decoded and those awaiting further decoding is given in Equation \ref{eqn:dec_memory} where $\alpha$ is the maximum active tokens after pruning (\textit{max-active}), $n_l$ is the maximum number of lanes, and $n_c$ is the maximum number of channels configured.
\begin{equation}
\label{eqn:dec_memory}
    M_{state} = 64\alpha n_c + 544\alpha n_l + 1024n_l
\end{equation}

Note that the size of the decoder state is \textit{not} related to the size of the decode graph nor the beam sizes.
As such, one can scale the decoder based on the desired number of parallel streams or sizes of the acoustic/language model.
As a concrete example, one could configure an edge device for a single stream ($\alpha=10000, n_c=1, n_l=1$) and use only 5.8MB of device memory, while a datacenter-class GPU might support 5000 simultaneous streams in realtime ($\alpha=10000, n_c=5000, n_l=500$) requiring about 5.5GB.

\subsection{Load Balancing}
To maximize parallelism, it is important that we generate large numbers of threads which have approximately the same amount of work to do. As we process each batch of frames, we begin by performing a load-balanced expand (see Figure \ref{fig:adv_dec}) where each outgoing arc is processed by its own thread, generating a number of candidate tokens. The adaptive beam is then adjusted, and used to determine which candidates are added back to the main queue for further processing.

Another irregularity comes from the slow convergence of non-emitting iterations, leading to an undefined number of small iterations (i.e. long tail). Once the count of active non-emitting tokens becomes low enough, the following iterations will be processed by a persistent kernel until convergence. In that persistent kernel, each utterance owns only one CUDA Cooperative Thread Array (CTA), speeding up synchronization and intra-thread communication.

\subsection{Lattice Preprocessing}
Up until the lattice processing stage in the decoder, the goal is to discover which subset of the search space would be saved for the current frame. Following frames build on that subset, and any paths within that subset may be present in the final lattice. During the discovery stage, we had to create and consider (typically an order of magnitude) more tokens than the ones we ultimately keep. Subsequently, the discovery stage focuses on being lightweight, while postponing any expensive structuring operations. 

In order to generate a lattice based on these tokens, we convert the raw tokens into a structured CSR representation. This includes detecting tokens linked to the same FST state, listing them in the CSR format, designing a unique representative for each FST state, and computing extra costs. This data is then moved to the host and used to generate the final lattice at the end of utterance. Tokens are then prepared for the next frame by ``soft-pruning'' any tokens which aren't representative for their FST state by artificially zeroing their out-arc degree, which can then be safely ignored by the load balancer: avoiding exponential growth.

\section{Experiments}
\label{sec:experiments}
We focus our examination on the performance of two models representing a wide spectrum of deployment conditions: from LibriSpeech \cite{librispeech} \textit{test-clean} subset evaluated with a model tuned specifically for LibriSpeech\footnote{Using standard Kaldi LibriSpeech recipe}, to the LibriSpeech \textit{test-other} subset evaluated on the ASPiRE \cite{aspire} Kaldi model\footnote{Available from http://kaldi-asr.org/models/m1}.
The former represents an ideal case of relatively easy-to-transcribe data being processed by a well-tuned model, while the latter is a more pathological case representing more challenging input audio transcribed by a mismatched model.
The net effect of the matched versus mismatched conditions is that in the case of the former, acoustic model posteriors tend to be more confident and fewer paths need to be evaluated when compared to more challenging scenarios.
All experiments are performed using a single NVIDIA Tesla V100 GPU, \textit{beam=15}, \textit{lattice-beam=8}, and \textit{max-active=10000}, unless otherwise specified.

\subsection{Accuracy}
The parallel implementation leads to expected non-determinism, typically due to out-of-order pruning of tokens. 
Specifically, the histogram pruning thresholds are somewhat arbitrary compared to the explicit cutoff in the baseline implementation.
Because of this, we see minor variations in the word error rate ($\pm 0.02\%$).
\begin{table}[thb]
\centering
\scalebox{0.75}{
\npdecimalsign{.}
\nprounddigits{0}
\begin{tabular}{l|ccc|ccc} 
 \hline
 \multirow{2}{*}{\textbf{Decoder}}    &
 \multicolumn{3}{c}{\textbf{test-clean}}  &
 \multicolumn{3}{c}{\textbf{test-other}} \\
  & \textbf{lat. den.}& \textbf{WER} & \textbf{OWER} & \textbf{lat. den.} & \textbf{WER} & \textbf{OWER}\\
 \hline
\textit{Baseline} & 4.19 & 5.49 & 1.05 & 13.94 & 13.71 & 2.55\\
GPU & 4.22 & 5.51 & 1.09 & 14.18 & 13.72 & 2.67 \\
 \hline
\end{tabular}
\npnoround
}

\caption{Validation of lattice quality with LibriSpeech model and test sets.}
\label{tab:correctness}
\end{table}

Table \ref{tab:correctness} evaluates the output lattices against lattices generated by the baseline CPU implementation. 
We validate that the word error rate (WER) is within tolerable limits, as well as the oracle word error rate (OWER). 
The Oracle WER is a proxy for determining if all expected alternate paths exist within the lattice. 
Finally, we measure the lattice density (\textit{lat. den.}), which is an average measure of outgoing arcs. This confirms the produced lattices are of similar size.

\subsection{Speed Improvements}
Table \ref{tab:rtf} reports xRTF (times faster than real time) for baseline Kaldi single- and multi-process decoder implementations, and other GPU decoder implementations.
The CPU speeds are obtained using an Intel Xeon CPU E5-2698 v4 @ 2.20GHz, with 20 cores.

Across the tested configurations, the GPU decoder outperforms the multithreaded CPU implementation within Kaldi, with a relative speedup ranging between 14x and 18x when compared to a full 20-core Xeon processor.
When compared with the current state-of-the art parallel decoder \cite{Chen2018}, the proposed algorithm decodes between 11x and 41x faster.
\begin{table}[thb]
\centering
\scalebox{0.75}{
\npdecimalsign{.}
\nprounddigits{1}
\begin{tabular}{ll|n{3}{2}n{3}{2}| n{4}{2}n{3}{2}} 
 \hline
 \multirow{2}{*}{\textbf{Decoder}}    &
 \multirow{2}{*}{\textbf{Type}}  &
 \multicolumn{2}{c}{\textbf{ASPiRE}}  &
 \multicolumn{2}{c}{\textbf{LibriSpeech}} \\
  && \textbf{clean}& \textbf{other} & \textbf{clean} & \textbf{other}\\
 \hline
CPU Process & One Best & 4.40 & 2.852 & 57.23966655 & 26.02413813\\
 \hline
CPU Process & Lattice & 3.785 & 2.7003 & 53.36967675 & 29.16244563\\
CPU Socket & Lattice & 43.24132673 & 30.11372356 & 614.8072061 & 313.1499666\\
 \hline
GPU \cite{Chen2018} & Lattice & 70.93265363 & \ \ \ \ n/a & 219.8842506 & 174.5955075 \\
GPU \textit{(This Work)} & Lattice &  {\npboldmath}769.260963 &  {\npboldmath}649.662766 & {\npboldmath}9031.446325 & {\npboldmath}4391.691471 \\
 \hline
\end{tabular}
\npnoround
}
\caption{Offline decoding speed (xRTF, \textit{beam=15})\tablefootnote{Missing data for prior GPU implementation is due to application crashes.}.}
\label{tab:rtf}
\end{table}

\subsection{Hyperparameters}
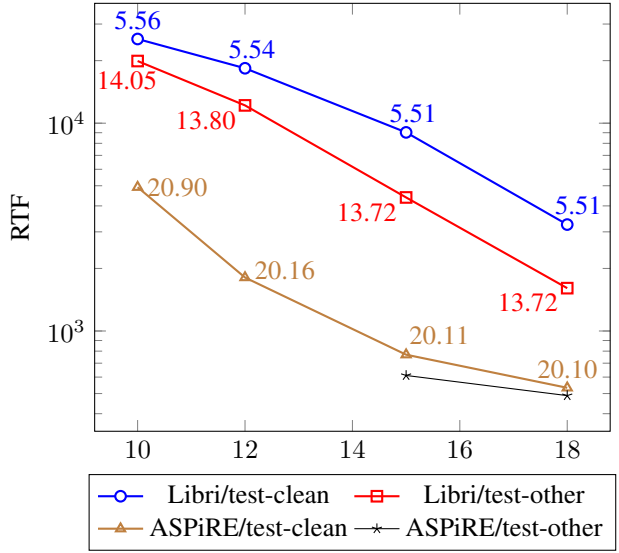
\begin{figure}
\begin{tikzpicture}
\begin{axis}[
	xlabel=Beam,
	ylabel=RTF,
	ymode=log,
	legend style={at={(0.5,-0.1)},anchor=north},
	legend columns=2]
\addplot[
    blue,
    thick,
    mark=*,
    mark options={fill=white},
    visualization depends on=\thisrow{alignment} \as \alignment,
    nodes near coords, %
    point meta=explicit symbolic, %
    every node near coord/.style={anchor=\alignment} %
] table [
    meta index=2
]{
x   y             label   alignment
10  25420.033959  5.56      270
12  18381.602773  5.54      250
15  9031.446325   5.51      250
18  3252.262073   5.51      240
};
\addplot[
    red,
    thick,
    mark=square,
    visualization depends on=\thisrow{alignment} \as \alignment,
    nodes near coords, %
    point meta=explicit symbolic, %
    every node near coord/.style={anchor=\alignment} %
] table [
    meta index=2
]{
x   y   label   alignment
10   19915.342576   14.05   60
12   12179.605955   13.80   20
15   4391.691471    13.72   20
18   1607.564912    13.72   20
};

\addplot[
    brown,
    thick,
    mark=triangle,
    visualization depends on=\thisrow{alignment} \as \alignment,
    nodes near coords, %
    point meta=explicit symbolic, %
    every node near coord/.style={anchor=\alignment} %
] table [
    meta index=2
]{
x   y   label   alignment
10   4920.583368    20.90   180
12   1808.737095    20.16   190
15   769.260963     20.11   210
18   532.972880     20.10   270
};

\addplot coordinates {
	(10,   0)
	(12,   0)
	(15,   611.004405)
	(18,   488.772100)
};

\legend{Libri/test-clean, Libri/test-other, ASPiRE/test-clean, ASPiRE/test-other}
\end{axis}
\end{tikzpicture}

\caption{RTF vs beam width.}
\label{fig:beam}
\end{figure}

Decoding hyperparameter selection (particularly \textit{beam}) impacts decoder speed. In cases with smaller beam widths, oversubscription of threads to the GPU is reduced, enabling faster inference. 
Care should be taken to choose a beam width that is suitable for the target data and model.
Figure \ref{fig:beam} shows a roughly log-linear decrease in decode speed as beam width increases.
The points in the graph are labeled with WER at that operating point.
Note the marginal accuracy improvements despite significant increases in runtime.

\begin{table}[thb]
\centering
\scalebox{0.75}{
\npdecimalsign{.}
\begin{tabular}{lc|n{2}{1}n{3}{1}| n{2}{1}n{3}{1}} 
 \hline
 \multirow{2}{*}{\textbf{LM}}    &
 \textbf{HCLG}    &
 \multicolumn{2}{c}{\textbf{test-clean}}  &
 \multicolumn{2}{c}{\textbf{test-other}} \\
  & \textbf{Size (MB)} & \textbf{xRTF} & \textbf{WER} & \textbf{xRTF} & \textbf{WER}\\
 \hline
3-gram, 3\text{e}{-10} & 192.6 &  5.51 & 9031.4 & 13.72 & 4391.7 \\
3-gram, 1\text{e}{-10} & 467.0 & 4.92 & 9064.5 & 12.54 & 4386.8 \\
3-gram & 8724.0 & {\npboldmath}4.02 & {\npboldmath}9161.7 & {\npboldmath}10.09 & {\npboldmath}4627.4 \\
 \hline
\end{tabular}
\npnoround
}

\caption{Comparison of FST size and WER/Speed.}
\label{tab:fst_size}
\end{table}
Table \ref{tab:fst_size} shows that significant reductions in WER may be achieved by using larger language models.
Three different trigram language models with different pruning thresholds ($3\text{e}{-10}$, $1\text{e}{-10}$, and no pruning, respectively) are used with other parameters held constant. Despite a 10x filesize difference, the decode performs \textit{faster} using the large language model likely due to reduced perplexity during decoding yielding extra pruning, and subsequently improved speed.

\subsection{Deployment}
With fully GPU-accelerated inference, the CPU is only left responsible for shuffling data in/out of the GPU, and completing lattice determinization if required.
Because of this, multi-GPU scaling is nearly linear.
On a NVIDIA DGX-1 containing 8 V100 GPUs, 85\% scaling efficiency is achieved when using all GPUs.

\begin{table}[thb]
\centering
\scalebox{0.75}{
\begin{tabular}{ll|c|c|c} 
 \hline
 \textbf{GPU}    &
 \textbf{Class}  &
 \textbf{Streams (10)}  &
 \textbf{Streams (15)}  &
 \textbf{TDP} \\
 \hline
Jetson Nano & Embedded & 11 & 7 & 5 \\
AGX Xavier & Embedded & 502 & 399 & 30 \\
Tesla T4 & Datacenter & 2024 & 1561 & 70 \\
Tesla V100 & Datacenter & 4117 & 3150 & 250 \\
 \hline
\end{tabular}
\npnoround
}

\caption{Measured end-to-end realtime throughput across suite of NVIDIA GPUs at varying beam sizes.}
\label{tab:e2ex}
\end{table}

Table \ref{tab:e2ex} demonstrates the same decoder used across the entire current NVIDIA family of processors.
In all cases, the models are identical, and use the same hyperparameters except for batch size.
The values in the table represent the number of streams that can be decoded in realtime, and includes feature extraction and acoustic model.

\section{Conclusion}
In this paper, we present a parallel decoder for speech recognition WFST inference. The algorithm is AM and LM agnostic, requiring no changes to support inference with existing models trained in the Kaldi toolkit. By implementing the decoder such that multiple utterances are processed in parallel, optimized memory management, and trading extra computation for reduced synchronization, we consistently achieve order-of-magnitude speedups when compared to the baseline multithreaded algorithm on CPU and current state-of-the-art GPU implementation. We further demonstrate that this work can be used on embedded platforms without requiring any model changes.

The implementation is now open-source as part of the Kaldi release. Future work will evaluate adaptations for CTC decoding as well as adding support for on-the-fly neural language model scoring.

\clearpage
\section{Bibliography}
\printbibliography[heading=none]

\end{document}